# SYNTACTIC ANALYSIS BY LOCAL GRAMMARS AUTOMATA: AN EFFICIENT ALGORITHM

Mehryar MOHRI
LADL-IGM[1]
Université Marne la Vallée
e-mail address: mohri@univ-mlv.fr

*Abstract*

The description of the constraints restricting words' combinations in specific contexts provides helpful grammars for reducing the number of ambiguities of lemmatized texts. These grammars allow to easily eliminate many of the ambiguities without even using complex general syntactic rules involving a lexicon-grammar. Local grammars can be represented in a very natural way by finite state automata. This paper describes and illustrates an efficient algorithm which allows to apply local grammars put in this shape to automata representing texts.

## 1. Introduction

One of the main incentives of syntactic analysis is to eliminate irrelevant ambiguities of a text. Local grammars constitute useful descriptions which help to remove some of these ambiguities. They consist of the description of local constraints, namely restrictions on the surrounding sequences of a given set of words. Combinations of French pre-verbal particles (see M. Gross 1989), in some extent agreement rules, other constraints independent of lexicon-grammar's entries, and many rules useful for error correction in texts provide typical examples of local grammars[2].

As shown further, local grammars can be represented in a very convenient way by finite state automata. The corresponding automata describe sets of locally unacceptable sequences[3] that a correct text should not contain. Once tagged, a text can in fact be itself represented by an automaton. Each of its path then constitutes an ambiguity. Thus, checking its correctness consists in removing the paths containing any of the forbidden sequences of the local grammar's automaton.

This requires searching in the automaton of the text for all occurrences of sequences of the local grammar's automaton. Such an operation can be considered as a generalization of the classical string matching problem which consists in finding all occurrences of a word in a text: here, we need to find a set of sequences in a set of texts. Several efficient algorithms have been proposed for solving the problem of locating occurrences of a finite set of sequences in a single text (A. V. Aho and M. J. Corasick 1975, B. Commentz-Walter 1979), and the application of local grammars to texts has already been described by E. Roche (1992).

---

[1] Laboratoire d'Automatique Documentaire et Linguistique and Institut Gaspard Monge.
[2] See M. Rimon and J. Herz (1991), and, F. C.N. Pereira and R. N. Wright (1991) for other related use of automata in syntactic analysis, and D. Maurel (1989) for a description of time expressions constraints in French by local grammars.
[3] Local grammars can be represented in an equivalent way by the set of obligatory sequences.

Here, we shall present a more efficient algorithm with a better time complexity which uses the notion of *failure function* or *default function* brought in by A. V. Aho and M. J. Corasick (1975) and extends it to the representation of automata. Analogous extensions have already been operated by M. Crochemore (1986). They show failure functions to be a helpful notion in the representation of automata.

In the following, we first illustrate the application of local grammars by considering several examples, then give a complete description of our algorithm and indicate corresponding experimental results.

**2. Application of local grammars**

Simple local rules can be easily represented by finite state automata. Consider, for instance, the word *this* in English. It is ambiguous for it can be a determiner, a demonstrative adjective, as in the following sentence:

>   *This program works well*,

a demonstrative pronoun as in:

>   *This does not change his opinion*,

or an adverb:

>   *He is not this tall*.

However, *this* imposes constraints to the choice of words it precedes. Simple observations lead to the following rules:

  *i*) when *this* is a determiner it cannot be followed by a verb unless the verb is a past or present participle;

  *ii*) the adverb *this* cannot be followed by a noun nor by a verb.

Rule *i*) can be illustrated by the following sentences:

>   **This sing is pretty*
>    *This falling rock is dangerous*
>    *He hates this inherited impatience of Lea*.

When *this* is an adverb, it can be followed by an adjective or by some other adverbs like *much* as in:

>   *He is not this much cleverer than her*[4].

Thus, the second rule indicates only some of the forbidden sequences in this case.

---

[4] Replacing *this* by *that* makes this sentence stylistically better.

Notice that the above rules are expressed in a negative way. Thus, it is quite natural to represent them by an automaton storing the set of unacceptable combinations. Figure 1 illustrates this automaton[5].

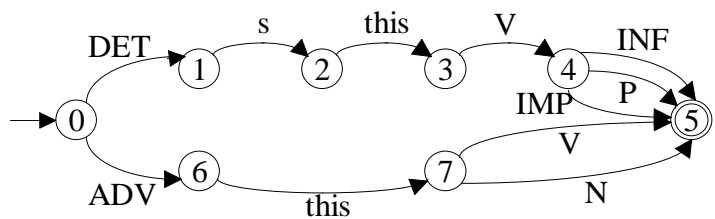

**Figure 1.** Local grammar of the form *this*.

It can be used to eliminate some of the ambiguities encountered in a text. Consider the sequence *this limit* which can be found in various contexts. *limit* is also an ambiguous form as it can be a verb conjugated at present at any person except the third of singular or an infinitive or an imperative form, or a singular noun. Thus, a simple dictionary look up allows to represent this sequence by the following automaton. Each path from the first state 0 to the final state 6 constitutes a possible analysis. However, the local grammar above helps to eliminate some of these ambiguities. Each path containing a sequence of the corresponding automaton can be removed.

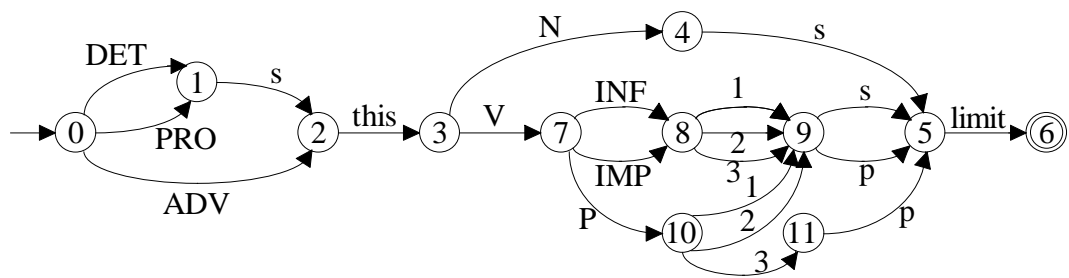

**Figure 2.** Automaton of the text *this limit*.

Thus, the application of the local grammar must lead to the following automaton.

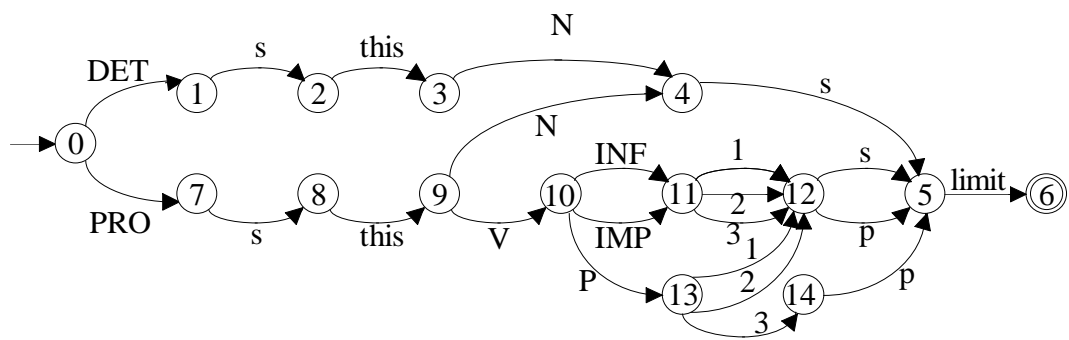

---

[5] Here, labels *p* and *s* stand for plural and singular, *ADV* for adverb, *DET* for determiner, *N* for noun, *PRO* for pronoun, *IMP* for imperative, *INF* for infinitive, *P* for present, *V* for verb and *1*, *2* and *3* indicate the first, second and third person. In order to simplify this presentation, we do not take into account the subjunctive here.

**Figure 3.** Automaton obtained after application of the local grammar.

Notice that many of the remaining paths can be part of acceptable sentences. The following sentences illustrate some of these possibilities:

*He made this limit the disaster*
*Let this limit his rudeness*
*After this, limit yourself to one per day.*

We shall describe in the next section the algorithm which can be used to perform the application of a local grammar when expressed in a negative way as above. However, in some cases, expressing rules in a positive way may be more appropriate. This can occur for example when acceptable sequences are less numerous or easier to describe than forbidden ones.

Agreement rules are often more easily expressed this way. Figure 4 gives a sample of agreement rules in French concerning articles *un* and *le* represented by an automaton. This automaton contains a set of obligatory sequences. It should be read in the following way: if *le* or *un* is a determiner masculine singular[6] followed by a noun, then this noun must also be masculine singular (paths '0 1 2 3 4 5'). Notice that this automaton gives no information about the constraints concerning a case where one of these determiner is followed by an adjective instead of a noun, and that it imposes no restriction on a sequence which does not contain these articles.

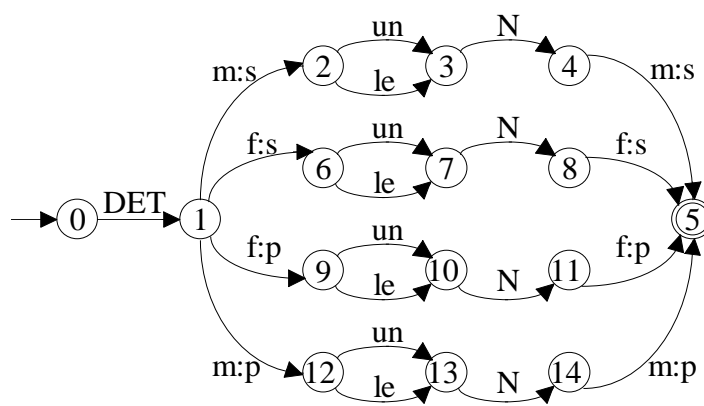

**Figure 4.** Agreement automaton for French articles *un* and *le*.

Such automata can also be used for disambiguation. In the next section we shall give more details about their definition and the corresponding algorithms[7].

## 3. Algorithm

We shall first consider the application of a local grammar represented by an automaton of forbidden sequences. Let $G_1 = (V_1, i_1, F_1, A^*, \delta_1)$ be a deterministic automaton representing the text, where $V_1$ is the set of its states, $i_1 \in V_1$ its initial state, $F_1 \subseteq V_1$ the set of its final states, $A$ its alphabet, and $\delta_1$ the state transition function which maps $V_1 \times A$ to $V_1$, and let $G_2 = (V_2, i_2, F_2, A^*, \delta_2)$ be the automaton representing the local grammar with analogous notations. In order to simplify the following algorithms we shall assume that $G_2$ is acyclic[8]. We denote by $L(G_1)$ (resp. $L(G_2)$) the language recognized by $G_1$ (resp. $G_2$). Thus, $A^*L(G_2)A^*$ constitutes the set of all sentences which contain an unacceptable sequence of $L(G_2)$. The application of the local grammar $G_2$ to $G_1$ should then lead to the regular language $L(G_1)\backslash A^*L(G_2)A^*$, namely the set of sentences of $L(G_1)$ which have no factor in $L(G_2)$. Here, we need to define an automaton $G = (V, i, F, A^*, \delta)$ recognizing this language[9].

---

[6] In the automata presented here we are only concerned with the canonical form of each word and with its morphological characteristics. Therefore, the feminine singular article *la* whose canonical form is *le* for instance is denoted by the sequence *DET f:s le*.

[7] Local constraints can also be represented by transducers (see M. Silberztein 1989). The minimization algorithm for transducers can help to limit the size of such transducers (see M. Mohri 1994). Here, we are only concerned with efficient algorithms involving automata.

In order to do so, we shall first indicate how to compute from $G_2$ a deterministic automaton representing the language $A^*L(G_2)$, namely the set of all sentences which end in $L(G_2)$.

### 3.1. Construction of a deterministic automaton recognizing $A^*L(G_2)$ from $G_2$

In general, constructing such an automaton is not a trite operation. It is easy to design a non-deterministic automaton recognizing $A^*L(G_2)$. Indeed, a simple loop labelled by all elements of the alphabet $A$ added at the initial state of $G_2$ is enough to transform it into an automaton recognizing $A^*L(G_2)$. The same can be done at the final states to obtain a non-deterministic automaton recognizing $A^*L(G_2)A^*$. However the use of this automaton makes the whole operation of application of the local grammar inefficient. Notice that the size of the alphabet $A$ is superior to the one of a dictionary of simple words of the language. Moreover, in some cases as in error correction applications the whole list of the elements of $A$ may not be available. Thus, a simple determinization of this automaton can be time consuming and even impossible in some case.

In order to construct a deterministic automaton recognizing $A^*L(G_2)$, we shall use the notion of failure function and gradually modify the automaton $G_2$. Consider a sequence $w \in A^*$. To know whether $w$ is in $A^*L(G_2)$, we can try to read it using the automaton $G_2$. As long as there is a transition corresponding to the read word in $G_2$ we use this transition to step to the following state. This allows us to read a prefix $x$ of $w$ in $G_2$. If the sequence $w$ is entirely read this way ($x = w$) and the reached state is a final state, then $w$ is in $L(G_2)$ and a fortiori in $A^*L(G_2)$. If not, then $w$ may have a suffix $v$ in $A^*L(G_2)$ (see figure 5). As shown by the figure below, $x$ has then a suffix $x'$ which is a prefix of $v$. In order to check the existence of a sequence like $v$, we need to start at a position as much at left as possible. In other words, we need to check whether $v$ is in $L(G_2)$ when $v$ is such that: $x'$ is the longest proper suffix of $x$ which is also a prefix of a sequence of $L(G_2)$.

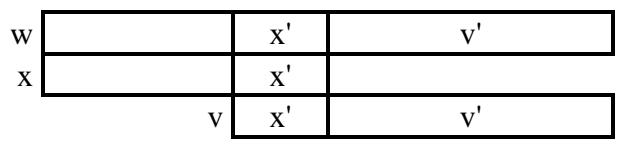

---

[8] The algorithms presented here can be easily modified in order to handle the case of non acyclic automata. However, local grammars are generally represented by acyclic automata. Besides, $G_1$, the text automaton, is of course also acyclic although we will not use this condition in the following.

[9] Notice the similarity of this problem with the one of string matching which consists of finding an occurrence of a word $x$ in a text $t$ and which can be expressed by: $t \in A^*xA^*$ ?

**Figure 5.** The definition of the failure function.

This leads to the definition of a function which associates to each state $u$ of $G_2$ the longest proper suffix of the paths reaching $u$ which is in $G_2$. However, this function can only be well-defined if all paths reaching $u$ have the same longest proper suffix $x'$ in $G_2$. Thus, in case two paths of $G_2$ leading to $u$ have different longest proper suffix $x'$ in $G_2$, we need to duplicate $u$ in two states corresponding to each of these paths. So, we can define a *failure function s*, which associates with each state $u$ the state corresponding to $x'$, namely $\delta_2(i_2, x')$. This function is to be consulted whenever the desired transition does not exist at a given state $u$ (A. V. Aho and M. J. Corasick 1975). Thanks to the use of a failure function, it is possible to represent the desired automaton even if the alphabet $A$ is infinite or undefined.

The following figures give an example of an automaton $G_2$ and its associated automaton $G_3$ which represents $A*L(G_2)$.

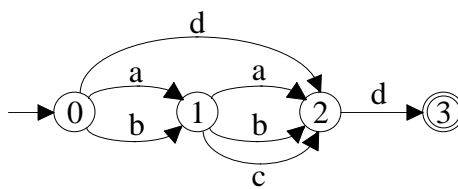

**Figure 6.** Local grammar $G_2$.

State numbers on the graph of figure 7 are followed by a slash and the value of the failure function at that state. For example, we have $s[4] = 1$, as the longest proper suffix of the sequence *aa* which is recognized by $G_2$ is *a*, the state corresponding to *a* is 1, and considering other combinations *ab*, *ba*, *bb* leads to the same result.

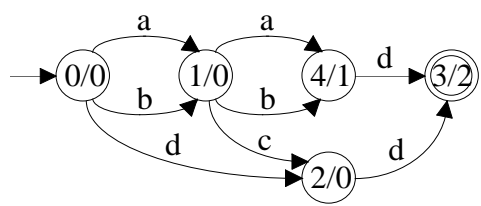

**Figure 7**. $G_3$, a deterministic automaton for $A*L(G_2)$.

Notice that the construction of this automaton has required the duplication of the state 2 of $G_2$, as *bb* and *bc* do not have the same longest proper suffix in $G_2$ (resp. *b*, and the empty word $\varepsilon$). The recognition process by such an automaton is quite easy. One just needs to use default transitions whenever usual transitions are not available. Consider for instance the sequence *aacd* which is in $A*L(G_2)$. The consecutive steps of the recognition of this sequence are:

    0  *aacd*
    1  *acd*
    4  *cd*
    1  *cd*   (failure transition)
    2  *d*
    3  $\varepsilon$.

As 3 is a final state, the sequence *aacd* is correctly recognized by $G_3=(V_3, i_3, F_3, A^*, \delta_3)$. It can be easily showed that the recognition process of a sequence *w* can still be done in $O(|w|)$ when using a representation by default functions.

The failure function *s* can be computed in an easy way. Indeed, it can be proved that for any state *u* and any element *a* of the alphabet *A*, $s[\delta(u, a)]$ is the first $\delta(s^k[u], a)$, $(k \geq 1)$, such that $\delta(s^k[u], a)$ is defined, or the initial state if none of these is defined. This gives a recursive algorithm for calculating this function. Notice that the definition of the failure function involves proper suffixes, hence for any state *u*, the level of the state $s[u]$ is lower than the one of *u*. In order to check whether a transition $\delta(s^k[u], a)$ is defined we then need to have defined and computed *s* for all states *v* with lower levels than *u*. This restricts the ordering in which the states should be considered in the algorithm. A breadth-first search (A. V. Aho *et al.* 1974, B. Sedgewick 1988, T. H. Cormen *et al.* 1990) of the automaton meets the corresponding condition. It suggests the use of a first-in first-out queue *Q* for managing the set of states to visit at each step. Figure 8 gives a pseudocode for an algorithm computing a deterministic automaton recognizing $A^*L(G_2)$ from $G_2$.

```
1   for each u ∈ V(G₂)
2       do  s[u] ← UNDEFINED
3   Q ← {i}
4   s[i] ← {i}
5   while Q ≠ ∅
6       do  u ← head[Q]
7           for each t ∈ Trans[u]
8               do  v ← s[u]
9                   while v ≠ i and δ(v, t.l) = UNDEFINED
10                      do v ← s[v]
11                  if u ≠ i and δ(v, t.l) ≠ UNDEFINED
12                      then  v ← δ(v, t.l)
13                  if s[t.v] = UNDEFINED
14                      then s[t.v] ← v
15                          ENQUEUE(Q, t.v)
16                          LIST-INSERT(list[t.v], t.v)
17                  else if there exists w ∈ list[t.v] such that s[w] = v
18                      then  t.v ← w
19                      else  w ← COPY-STATE(t.v)  ◊copy of t.v with same transitions
20                          s[w] ← v
21                          LIST-INSERT(list[t.v],w)
22                          t.v ← w
23                          ENQUEUE(Q, w)
24          DEQUEUE(Q)
```

**Figure 8**. Algorithm for the construction from $G_2$ of a deterministic automaton for $A^*L(G_2)$.

We here denote by *Trans*[*u*] the set of transitions leaving a state $u \in V$, and for each *t* in *Trans*[*u*] and $u \in V$, by *t.v* the vertex reached by *t* and *t.l* its label. We also use a special constant UNDEFINED different from all states of $G_2$. The algorithm directly modifies the automaton $G_2$ into one representing $A^*L(G_2)$, by duplicating states whenever it is necessary (function COPY-STATE), and by computing the failure function $s$ for all states. In order to limit the duplication of states, the list of copied states of a state $u$ is stored at each step in $list[u]$ and a new state is created only if no other equivalent state with the same default state exists.

Notice that the loop of lines 5-23 iterates as long as there remains a state $u$ for which all leaving transitions have not yet been examined. Each state $u$ is enqueued exactly once in $Q$. Hence, the total number of iterations of the loop 5-23 is equal to the number of states of the resulting automaton $G_3$. The loop of lines 7-22 is performed once for each transition leaving $u$. Thus, if we denote by $E(G_3)$ the set of the transitions of $G_3$, in the whole this loop is iterated $|E(G_3)|$ times. If the test made at line 17 and the insertion of line 16 are efficiently implemented by using hashing method, each iteration can be assumed to be done in constant time. The total running time of the algorithm including the initialization (lines 1-2) is then $O(|V(G_3)| + |E(G_3)|)$, thus linear in the number of states and transitions of the resulting automaton.

It is also worthwhile to point out that if the automaton $G_2$ is minimal[10], then the resulting automaton $G_3$ is also the minimal deterministic automaton representing $A^*L(G_2)$. Indeed, if two states $u$ and $u'$ were equivalent in $G_3$, then by definition[11] they would be copies of a same state $v$ of $G_2$. As states of $G_2$ are duplicated only if necessary, then $u$ and $u'$ bear different failure function's values. Therefore, different sequences can be read from $u$ and $u'$ to a final state of $G_3$. This contradicts the equivalence of these states.

There are some particular cases in which the size of the obtained automaton $G_3$ is exponential. The minimal automaton associated to $a(a + b)^n$ for instance has $(n + 2)$ states whereas it is easy to show that the minimal automaton of $(a + b)^*a(a + b)^n$ has $2^{n+1}$ states. However, such blow up cases are generally not encountered in Natural Language problems, and if they could occur then the result of the application of the corresponding local grammar could also have an exponential size.

In the following section, we shall indicate how to use the obtained automaton recognizing $A^*L(G_2)$ so as to apply the local grammar.

### 3.2. Application of the local grammar and experimental results

Once the automaton $G_3$ representing $A^*L(G_2)$ is provided, the application of the local grammar $G_2$ becomes considerably easier. Given an automaton $G_1$ representing a text, one can directly construct a deterministic automaton corresponding to the language $L(G_1)\backslash A^*L(G_2)A^*$, by using $G_1$ and $G_3$. Indeed, we can simultaneously read these two automata, store at each step the two states reached in each of them and keep those transitions of $G_1$ which do not lead to a final state of $G_3$. This can be illustrated by the following figures. Figure 9 gives an example of a text automaton, and figure 10 the automaton $G_4$ obtained by using $G_3$ (figure 7). Each state of $G_4$ bears a pair of numbers indicating the states reached respectively in $G_1$ and $G_3$.

---

[10] Notice that minimal automata representing local grammars of unacceptable sequences have only one final state, as there is no need adding an unacceptable sequence to such local grammars when a prefix of it is already part of forbidden sequences.

[11] Notice that if we do not take into account default transitions, then $G_3$ represents the same language as $G_2$.

The initial state of $G_4$ corresponds to the pair (0,0) of initial states of $G_1$ and $G_3$. Transitions $a$ and $b$, for instance both lead from this state to (1,1) as reading these transitions lead to state 1 in $G_1$ and also 1 in $G_3$.

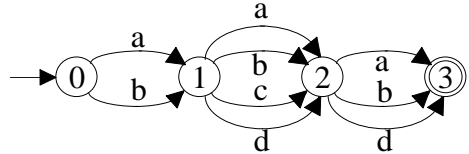

**Figure 9**. Text automaton $G_1$.

The transition labelled by $a$ from (1,1) leads to (2,4) as $\delta_1(1, a)=2$ and $\delta_3(1, a)=4$. Transitions by $d$ from (2,4) and (2,2) are not kept (represented by dotted line) as they lead to the final state 3 of $G_3$.

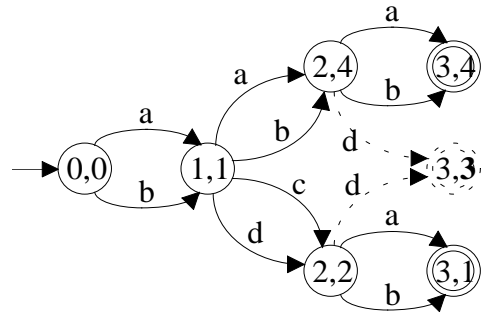

**Figure 10**. Automaton $G_4$ obtained from $G_1$ by application of the local grammar $G_2$.

This construction is similar to the one used to obtain the intersection of two automata.

LOCAL-GRAMMAR(G1, G3, G4)
1    $F_4 \leftarrow \emptyset$
2    $\{i_4\} \leftarrow (i_1, i_3)$
3    $Q \leftarrow \{i_4\}$
4    **while** $Q \neq \emptyset$
5      **do** $u_4=(u_1, u_3) \leftarrow head[Q]$
6        **for** each $t \in Trans[u_1]$   ◊  transitions considered in $G_1$
7          **do** $v_1 \leftarrow \delta_1(u_1, t.l)$
8            $v_3 \leftarrow u_3$
9            **while** $v_3 \neq i_3$ **and** $\delta_3(v_3, t.l) =$ UNDEFINED
10              **do** $v_3 \leftarrow s[v_3]$
11            **if** $\delta_3(v_3, t.l) \neq$ UNDEFINED
12              **then** $v_3 \leftarrow \delta_3(v_3, t.l)$
13            **if** $v_3 \notin F_3$
14              **then** $v_4 \leftarrow (v_1, v_3)$

```
15                          if v₄ is a new state
16                             then ENQUEUE(Q, v₄)
17                                  if v₁ ∈ F₁
18                                     then F₄ ← F₄ ∪ {v₄}
19                          δ₄(u₄, t.l) ← v₄
20              DEQUEUE(Q)
```

**Figure 11**. Algorithm for the application of a local grammar.

The simple pseudocode above gives the algorithm computing $G_4$. This algorithm is efficient as it does not require to inspect transitions leaving a set of states at each step of its execution but only those corresponding to a pair of states, one in $G_1$ and one in $G_3$. In case the test performed at line 15 is considered to be performed in constant time[12], the algorithm can be showed to be quadratic, more precisely in $O(|V(G_3)|.(|V(G_1)| + |E(G_1)|))$.

We have implemented and experimented this algorithm and the one presented in the previous section. We have tested these algorithms by considering a set of 1.600 sequences of length 20 or more. The corresponding minimal automaton had about 18.000 states. We then defined a simple automaton of about 290 states so as to simulate a local grammar[13].

The first algorithm applied to this automaton led to an automaton of about 340 states. We have checked the fact that the number of states of the automaton does not grow exponentially after application of this algorithm by carrying on several experiments with automata reaching the size of about 2500 states. In our experiments, the number of states of the resulting automaton never exceeded one and a half the one of the initial automaton. The time spent for the execution of this algorithm never reached the second[14].

Notice that once the operation corresponding to this algorithm has been performed the resulting automaton can be used for disambiguating any text. Hence, the time spent for the construction of this automaton can be considered as pre-processing of the grammar and done once for all. The two algorithms described can be combined into a single one such that only necessary states of the automaton $G_2$ be considered and that only corresponding failure function values be evaluated. The previous remark, however, reduces the interest of such an algorithm.

The application of the local grammar to the text following the algorithm above results in an automaton which had about 17.300 states. More precisely, this figure corresponds to the automaton obtained after minimization. The whole process of application of the local grammar, including this minimization, took about 10'. This suggests that the algorithm presented above is very efficient for the sizes we considered.

### 3.3.   Local constraints described in a positive way

---

[12] This is roughly the case if we use an efficient hashing method for the implementation.

[13] The number of states of the local grammars we had at our disposal did not exceed fifty. This simulation aimed at anticipating the fast growth of creation or definition of new local grammars. Their union could very soon reach several hundreds of states.

[14] Our program was written in C and implemented on a Next Cube, processor 68040 33Mhz with 32 Mb of RAM, and on a IBM PS/2 i486 50 Mhz with 16 Mb of RAM.

We have already indicated the possibility of representing the set of local obligatory sequences by automata. Here we shall give more details about their exploitation.

Positive rules are often expressed in the following way: if an expression contains the sequence *X*, then it must be followed by the word *y*, or the last word of *X* must have the property *z*. This appears clearly in agreement rules as in the example indicated for French articles followed by a noun. Hence, the paths of an automaton corresponding to positive rules do not necessarily constitute all possible paths in a given context. They can only be used in the following way: if the beginning *X* of one of these paths is encountered in a text and if *Xy* leads to a final state, then the corresponding part of the text must also be followed by the label *y* or another label *y'* such that *Xy'* be the beginning of a path in the local grammar automaton. Thus, final states play an important role in positive local grammar automata.

The automaton below represents a part of the agreements on gender and number of the French article *un* with the adjectives or nouns following it. The notation '?' is here used to represent any possible canonical form in this context, and *A* stands for adjective.

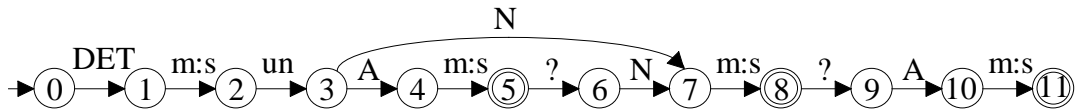

**Figure 12**. Part of agreements on gender and number of article *un* and nouns and adjectives.

It can be read in the following way: if *un* is a determiner masculine singular followed by an adjective, then this adjective is also masculine singular, etc. This allows a natural positive representation of some local constraints. Indeed, here, one only needs to describe as much as possible the context of a given sequence, and then impose corresponding constraints by using final states.

POSITIVE-LOCAL-GRAMMAR(G1, G3, G4)
1    $F_4 \leftarrow \emptyset$
2    $\{i_4\} \leftarrow (i_1, i_3)$
3    $Q \leftarrow \{i_4\}$
4    **while** $Q \neq \emptyset$
5        **do** $u_4 = (u_1, u_3) \leftarrow head[Q]$
6            **for** each $t \in Trans[u_1]$   ◊   transitions considered in $G_1$
8                **do** $v_3 \leftarrow u_3$
7                    **while** $v_3 \neq i_3$ **and** $\delta_3(v_3, t.l) =$ UNDEFINED **and** $FT[v_3]$=TRUE
8                        **do** $v_3 \leftarrow s[v_3]$
9                    **if** $\delta_3(v_3, t.l) \neq$ UNDEFINED **or** ($v_3 = i_3$ **and** $FT[v_3]$=FALSE)
10                        **then** $v_1 \leftarrow \delta_1(u_1, t.l)$
11                            **if** $\delta_3(v_3, t.l) \neq$ UNDEFINED
12                                **then** $v_3 \leftarrow \delta_3(v_3, t.l)$
13                            $v_4 \leftarrow (v_1, v_3)$
14                            **if** $v_4$ is a new state

```
15                              then ENQUEUE(Q, v₄)
16                                 if v₁ ∈ F₁
17                                    then F₄ ← F₄ ∪ {v₄}
18                       δ₄(u₄, t.l) ← v₄
19         DEQUEUE(Q)
```

**Figure 11**. Algorithm for the application of a positive local grammar.

The application of such automata is close to the one described previously. Here too, we shall use the first algorithm presented above in order to construct an automaton $G_3$ recognizing $A*L(G_2)$ from $G_2$. Only the application of $G_3$ slightly differs from the one indicated above. Instead of keeping those transitions of $G_1$ which do not lead to a final state of $G_3$, here we shall reject only those which do not exist in $G_3$ whereas this graph contains another transition leading to a final state. The corresponding algorithm can be obtained easily from the one indicated above. Figure 11 describes this algorithm. Here, we have denoted by $FT[v_3]$ the following property: none of the adjacent state of $v_3$ is a final state ($\forall\ t' \in Trans[u_3], t'.v \notin F_3$).

This algorithm has obviously the same complexity as the one presented above. Therefore, the use of negative or positive rules in the representation of local grammars has no algorithmic effect on their application, and the choice of the appropriate representation should be mainly motivated by practical or heuristical considerations which can allow, in particular, to reduce the size of the involved automata.

## 4. Conclusion

The number of local grammars allowing to represent more conveniently contextual constraints keeps increasing. Hence, so does the size of the union of all the corresponding automata. The algorithms described here should allow to apply efficiently such automata even with large sizes in order to reduce the number of ambiguities of texts. They also make it more natural to use automata to impose constraints on factors of a text.

Many other operations related to syntactic analysis by automata such as intersections of the form $(A*L(G)A* \cap G')$ involve the computation for a given automaton $G$ of a deterministic one representing $A*L(G)$. The presented algorithms can also improve the efficiency of these operations. They can also be used in other applications such as pattern matching when the provided data is not a list of words to search for in a text, but an automaton representing these words.